\documentclass[letterpaper]{llncs}{}
\usepackage{lineno,hyperref}
\usepackage[export]{adjustbox}
\usepackage{amssymb}
\usepackage{multicol}
\usepackage{graphicx} 
\usepackage{filecontents}
\usepackage{bbold}
\usepackage{makeidx}  
\usepackage[misc]{ifsym}
\usepackage{amsmath,graphicx}
\usepackage{multirow}
\usepackage{array}
\usepackage{siunitx}
\DeclareMathOperator*{\argmin}{arg\,min}
\DeclareMathOperator{\Tr}{Tr}
\newcommand{\rpm}{\raisebox{.2ex}{$\scriptstyle\pm$}}
\modulolinenumbers[5]
\usepackage{mathtools}

\usepackage{amsmath}
\usepackage{graphicx}
\usepackage{wrapfig}
\begin{document}
\title{A Coupled Manifold Optimization Framework to Jointly Model the Functional Connectomics and Behavioral Data Spaces}
%
%
\author{Niharika Shimona D'Souza\inst{1} 
 \textsuperscript{*}
\and Mary Beth Nebel \inst{2} \inst{3} \and Nicholas Wymbs \inst{2} \inst{3} \and Stewart Mostofsky \inst{2}\inst{3} \inst{4} \and Archana Venkataraman\inst{1} }
%

\authorrunning{Niharika Shimona D'Souza et al.} 
%
%

\institute{Dept. of Electrical and Computer Eng., Johns Hopkins University, Baltimore, USA \email{\textsuperscript{*}}{Shimona.Niharika.Dsouza@jhu.edu}
\and
Center for Neurodevelopmental Medicine \& Research, Kennedy Krieger Institute
\and
Dept. of Neurology, Johns Hopkins School of Medicine, Baltimore, USA
\and
Dept. of Pediatrics, Johns Hopkins School of Medicine, Baltimore, USA}
\maketitle              
\begin{abstract}
The problem of linking functional connectomics to behavior is extremely challenging due to the complex  interactions between the two distinct, but related, data domains. We propose a coupled manifold optimization framework which projects fMRI data onto a low dimensional matrix manifold common to the cohort. The patient specific loadings simultaneously map onto a behavioral measure of interest via a second, non-linear, manifold. By leveraging the kernel trick, we can optimize over a potentially infinite dimensional space without explicitly computing the embeddings. As opposed to conventional manifold learning, which assumes a fixed input representation, our framework directly optimizes for embedding directions that predict behavior. Our optimization algorithm combines proximal gradient descent with the trust region method, which has good convergence guarantees. We validate our framework on resting state fMRI from fifty-eight patients with Autism Spectrum Disorder using three distinct measures of clinical severity. Our method outperforms traditional representation learning techniques in a cross validated setting, thus demonstrating the predictive power of our coupled objective.
\end{abstract}
\section{Introduction}

\par Steady state patterns of co-activity in resting state fMRI (rs-fMRI) are believed to reflect the intrinsic functional connectivity between brain regions \cite{fox2007spontaneous}. Hence, there is increasing interest to use rs-fMRI as a diagnostic tool for studying neurological disorders such as autism, schizophrenia and ADHD. Unfortunately, the well reported confounds of rs-fMRI, coupled with patient heterogeneity makes the task of jointly analyzing rs-fMRI and behavior extremely challenging.
\paragraph{\textbf{Behavioral Prediction from Neuroimaging Data.}} Joint analysis of rs-fMRI and behavioral data typically follows a two stage pipeline. Stage $1$ is a feature selection or a representation learning step, while Stage $2$ maps the learned features onto behavioral data through a statistical or machine learning model. Some notable examples of the Stage 1 feature extraction include graph theoretic measures which aggregate the associative relationships in the connectome, and dimensionality reduction techniques \cite{murphy2012machine}, which explain the variation in the data. From here, popular Stage $2$ algorithms include Support Vector Machine (SVMs), kernel ridge regression \cite{murphy2012machine}. This pipelined approach has been successful at classification for identifying disease subtypes and distinguishing between patients and healthy controls. However, there has been limited success in terms of predicting dimensional measures, such as behavioral severity from neuroimaging data.
\par The work of \cite{d2018generative} develops a generative-discriminative basis learning framework, which decomposes the rs-fMRI correlation matrices into a group and patient level term. The authors use a linear regression to estimate clinical severity from the patient representation, and jointly optimize the group average, patient coefficients, and regression weights. In this work, we pose the problem of combining the neuroimaging and behavioral data spaces as a dual manifold optimization. Namely, we represent the each patient's fMRI data using a low rank matrix decomposition to project it onto a common vector space. The projection loadings are simultaneously used to construct a high dimensional non-linear embedding to predict a behavioral manifestation. We jointly optimize both representations in order to capture the complex relationship between the two domains. 
\paragraph{\textbf{Manifold Learning for Connectomics.}} Numerous manifold learning approaches have been employed to study complex brain topologies, especially in the context of disease classification. For example, the work of  \cite{thiagarajan2014multiple} used graph kernels on the spatio-temporal fMRI time series dynamics to distinguish between the autistic and healthy groups. Going one step further, \cite{soussia2017high} used higher order morphological kernels to classify ASD subpopulations.
\par While these methods are computationally efficient and simple in formulation, their generalization power is limited by the input data features. Often, subtle individual level changes are overwhelmed by group level confounds. We integrate the feature learning step directly into our framework by  simultaneously optimizing both the embeddings and the projection onto the behavioral space. This optimization is also coupled to the brain basis, which helps us model the behavioral and neuroimaging data space jointly, and reliably capture individual variability. We leverage the kernel trick to provide both the representational flexibility and computational tractability to outperform a variety of baselines.

\section{A Coupled Manifold Optimization (CMO) Framework}
\begin{figure}[t!]
   \centering
   \includegraphics[scale=0.54]{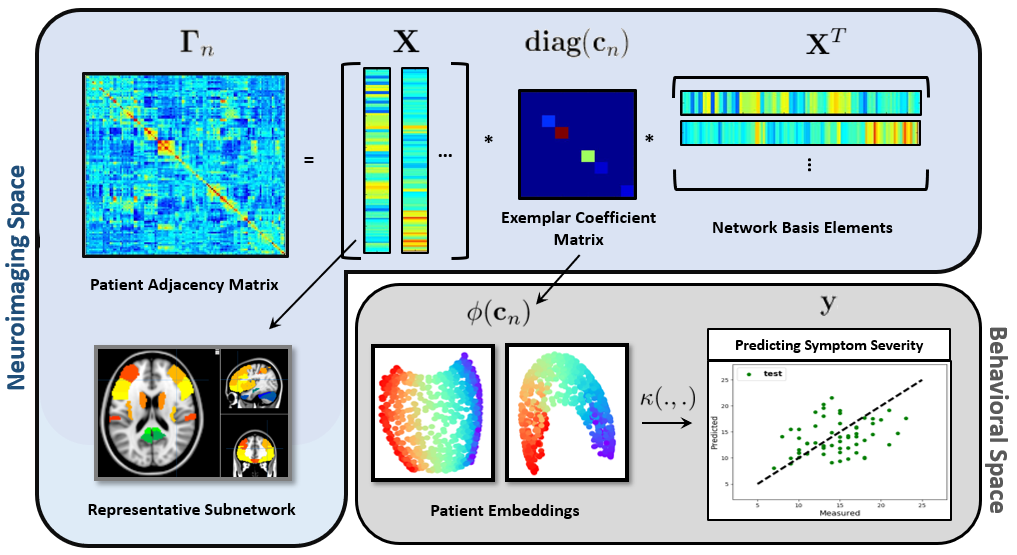}
   \small{\caption{\footnotesize{Joint Model for the Functional Connectomics and Behavioral Data. \textbf{Blue Box:}~Matrix Manifold Representation \textbf{Gray Box:} Non-Linear kernel Ridge Regression} } \label{CCS}}
\end{figure}
Fig.~\ref{CCS} presents an overview of our Coupled Manifold Optimization (CMO) framework. The blue box represents our neuroimaging term. We group voxels into $P$ ROIs, yielding the $P\times P$ input correlation matrices $\{\mathbf{\Gamma}_{n}\}_{n=1}^{N}$ for $N$ patients. As seen, the correlation matrices are projected onto a low rank subspace spanned by the group basis. The loadings are related to severity via a non-linear manifold and the associated kernel map, as indicated in the gray box.
\par Notice that $\mathbf{\Gamma}_{n}$ is positive semi-definite by construction. We employ a patient specific low rank decomposition $\mathbf{\Gamma}_{n} \approx \mathbf{Q}_{n} \mathbf{Q}^{T}_{n}$ to represent the correlation matrix. Each rank $R$ factor $\{\mathbf{Q}_{n} \in \mathcal{R}^{P \times R}\}$  , where $R \ll P$ , projects onto a low dimensional subspace spanned by the columns of a group basis $\mathbf{X} \in \mathcal{R}^{P \times R}$. The vector $\mathbf{c}_{n} \in \mathcal{R}^{R \times 1}$ denotes the patient specific loading coefficients as follows:
\begin{equation}
\mathbf{\Gamma}_{n} \approx \mathbf{Q}_{n} \mathbf{Q}_{n}^{T} = \mathbf{X}\mathbf{diag}(\mathbf{c}_{n}) \mathbf{X}^{T}
\label{Repres2}
\end{equation}
where $\mathbf{diag}(\mathbf{c}_{n})$ is a matrix with the entries of $\mathbf{c}_{n}$ on the leading diagonal, and the off-diagonal elements as $0$. Eq.~(\ref{Repres2}) resembles a joint eigenvalue decomposition for the set $\{\mathbf{\Gamma}_{n}\}$ and was also used in \cite{d2018generative}. The bases $\mathbf{X}_{r} \in \mathcal{R}^{P \times 1}$ capture co-activation patterns common to the group, while the coefficient loadings $\mathbf{c}_{nr}$ capture the strength of basis column~$r$ for patient~$n$. Our key innovation is to use these coefficients to predict clinical severity via a non-linear manifold. We define an embedding map $\mathbf{\phi}(\cdot):~ \mathcal{R}^{R} \rightarrow \mathcal{R}^{M} $, which maps the native space representation of the coefficient vector $\mathbf{c}$ to an $M$ dimensional embedding space, i.e. $\mathbf{\phi}(\mathbf{c}) \in \mathcal{R}^{M \times 1}$. If $\mathbf{y}_{n}$ is the clinical score for patient~$n$, we have the non-linear regression:
\begin{equation}
    \mathbf{y}_{n} \approx \mathbf{\phi}(\mathbf{c}_{n})^{T}\mathbf{w}
    \label{Reg}
\end{equation}
with weight vector $\mathbf{w} \in \mathcal{R}^{M \times 1}$.
Our joint objective combines Eq.~(\ref{Repres2}) and Eq.~(\ref{Reg})
\begin{equation}
\mathcal{J}(\mathbf{X},\{\mathbf{c}_{n}\},\mathbf{w}) = \sum_{n}{\left[\vert\vert{\mathbf{\Gamma}_{n} - \mathbf{X}\mathbf{diag}(\mathbf{c}_{n}) \mathbf{X}^{T}\vert\vert}_{F}^{2} + \lambda{\vert\vert{\mathbf{y}_{n} - \mathbf{\phi}(\mathbf{c}_{n})^{T}\mathbf{w}}\vert\vert}_{2}^{2}\right]}
\label{obj}
\end{equation}
along with the constraint $\mathbf{c}_{nr}\geq 0$ to maintain positive semi-definiteness of $\{\mathbf{\Gamma}_{n}\}$. Here, $\lambda$ controls the trade-off between the two representations. We include an $\ell_{1}$ penalty on $\mathbf{X}$ to promote sparse solutions for the basis. We also regularize both the coefficients $\{\mathbf{c}_{n}\}$ and the regression weights $\mathbf{w}$ with $\ell_{2}$ penalties to ensure that the objective is well posed. We add the terms $\gamma_{1}{\vert\vert{\mathbf{X}}\vert\vert}_{1} + \gamma_{2}{\sum_{n}{\vert\vert{\mathbf{c}_{n}\vert\vert}}^{2}_{2}} + \gamma_{3}{\vert\vert{\mathbf{w}}\vert\vert}^{2}_{2}$ to $\mathcal{J}(\cdot)$ in Eq.~(\ref{obj}) with the penalties $\gamma_{1}$,$\gamma_{2}$ and $\gamma_{3}$ respectively.
\subsection{Inferring the latent variables:} 
We use alternating minimization to estimate the hidden variables $\{\mathbf{X},\{\mathbf{c}_{n}\},\mathbf{w}\}$. This procedure iteratively optimizes each unknown variable in Eq.~(\ref{obj}) by holding the others constant until global convergence is reached.
\par Proximal gradient descent \cite{parikh2014proximal} is an efficient algorithm which provides good convergence guarantees for the non-differentiable $\ell_{1}$ penalty on $\mathbf{X}$. However, it requires the objective to be convex in $\mathbf{X}$, which is not the case due to the bi-quadratic Frobenius norm expansion in Eq.~(\ref{CCS}). Hence, we introduce $N$ constraints of the form $\mathbf{V}_{n} = \mathbf{X}\mathbf{diag}(\mathbf{c}_{n})$, similar to the work of \cite{d2018generative}. We enforce these constraints using the Augmented Lagrangians $\{\mathbf{\Lambda}_{n}\}$:
\begin{multline}
\mathcal{J}(\mathbf{X}, \{\mathbf{c}_{n}\},\mathbf{w},\{\mathbf{V}_{n}\},\{\mathbf{\Lambda}_{n}\}) = {\sum_{n}}{\vert\vert{\mathbf{\Gamma}_{n}-\mathbf{V}_{n}\mathbf{X}^{T}}\vert\vert}_{F}^{2}  + \lambda{\sum_{n}}{\vert\vert{\mathbf{y}_{n}-\mathbf{\phi}({\mathbf{c}_{n})^{T}\mathbf{w}}\vert\vert}_{2}^{2}} \\ + \sum_{n}{\left[{\Tr{\left[{\mathbf{\Lambda}_{n}^{T}({\mathbf{V}_{n}-\mathbf{X}\mathbf{diag}(\mathbf{c}_{n})})}\right]}} + {{\frac{1}{2}}{\vert\vert{\mathbf{V}_{n}-\mathbf{X}\mathbf{diag}(\mathbf{c}_{n})}\vert\vert}_{F}^{2}}\right]}
\label{obj2}
\end{multline}
with $ \mathbf{c}_{nr} \geq 0$ and $\Tr({\mathbf{M}})$ denoting the trace operator. The additional terms ${\vert\vert{\mathbf{V}_{n}-\mathbf{X}\mathbf{diag}(\mathbf{c}_{n})}\vert\vert}_{F}^{2}$ regularize the trace constraints. Eq.~(\ref{obj2}) is now convex in both $\mathbf{X}$ and the set $\{\mathbf{V}_{n}\}$, which allows us to optimize them via standard procedures. We iterate through the following four update steps till global convergence:

\paragraph{\textbf{Proximal Gradient Descent on $\mathbf{X}$:}}
The gradient of $\mathcal{J}$ with respect to $\mathbf{X}$ is:
\begin{equation*}
\frac{\partial\mathcal{J}}{\partial\mathbf{X}} = \sum_{n}{2\left[{\mathbf{X}\mathbf{V}_{n}^{T}-\mathbf{\Gamma}_{n}}\right]\mathbf{V}_{n}-\mathbf{V}_{n}\textbf{diag}(\mathbf{c}_{n})} +{\mathbf{X}\textbf{diag}(\mathbf{c}_{n})^{2}-\mathbf{\Lambda}_{n}\textbf{diag}(\mathbf{c}_{n})}
\end{equation*}
With a learning rate of $t$, the proximal update with respect to $\vert\vert{\mathbf{X}}\vert\vert_{1}$ is given by:
\begin{multline*}
\mathbf{X}^{k} =\mathbf{prox}_{{\vert\vert{\mathbf{\cdot}}\vert\vert_{1}}}\left[\mathbf{X}^{k-1} - \left[\frac{t}{\gamma_{1}}\right]\frac{\partial\mathcal{J}}{\partial\mathbf{X}}\right] \  s.t. \   \mathbf{prox}_{t}(\mathbf{L}) =
\mathbf{sgn}(\mathbf{L})\circ(\mathbf{max}(\vert{\mathbf{L}}\vert-t,\mathbf{0}))
\label{ProxUpd}
\end{multline*}
Where $\circ$ denotes the Hadamard product. Effectively, this update performs an iterative shrinkage thresholding on a locally smooth quadratic model of ${\vert\vert{\mathbf{X}}\vert\vert}_{1}$.

\paragraph{\textbf{Kernel Ridge Regression for $\mathbf{w}$:}} We denote $\mathbf{y}$ as the vector of the clinical severity scores and stack the patient embedding vectors i.e. $\mathbf{\phi(\mathbf{c}_{n})} \in \mathcal{R}^{M \times 1}$ into a matrix $\mathbf{\Phi(\mathbf{C})} \in \mathcal{R}^{M \times N}$. The portion of $\mathcal{J}(\cdot)$ that depends on $\mathbf{w}$ is:
\begin{equation}
\mathcal{F}(\mathbf{w}) =  \lambda{\vert\vert{\mathbf{y}-\mathbf{\Phi}(\mathbf{C})^{T}\mathbf{w}}\vert\vert}_{2}^{2} + \gamma_{3}{\vert\vert{\mathbf{w}}\vert\vert}^{2}_{2}  
\label{krr}
\end{equation}
Setting the gradient of Eq.~(\ref{krr}) to $0$, and applying the matrix inversion lemma, the closed form solution for $\mathbf{w}$ is similar to kernel ridge regression: 
\begin{equation}
\mathbf{w} = \mathbf{\Phi}(\mathbf{C}){\left[\mathbf{\Phi}(\mathbf{C})^{T}\mathbf{\Phi}(\mathbf{C})+ \frac{\gamma_{3}}{\lambda} \mathcal{I}_{N}\right]}^{-1} \mathbf{y} = \mathbf{\Phi}(\mathbf{C}) \boldsymbol{\alpha} = \sum_{j}{\boldsymbol{\alpha}_{j}{\mathbf{\phi}(\mathbf{c}_{j})}}
    \label{spaceproj}
\end{equation}
where $\mathcal{I}_{N}$ is the identity matrix. Let $ \kappa(\cdot,\cdot):~\mathcal{R}^{M} \times \mathcal{R}^{M} \rightarrow \mathcal{R}$ be the kernel map for $\mathbf{\phi}$, i.e. $\kappa(\mathbf{c},\hat{\mathbf{c}}) = \mathbf{\phi(\mathbf{c})^{T}} \mathbf{\phi(\hat{\mathbf{c}})}$. The dual variable $\boldsymbol{\alpha}$ can be expressed as $\boldsymbol{\alpha} = (\mathbf{K}+\frac{\gamma_{3}}{\lambda}\mathcal{I}_{N})^{-1}\mathbf{y}$, where $\mathbf{K} = \mathbf{\Phi}(\mathbf{C})^{T}\mathbf{\Phi}(\mathbf{C})$ is the Gram matrix for the kernel $\mathbf{\kappa}(\cdot,\cdot)$. Eq.~(\ref{spaceproj}) implies that $\mathbf{w}$ lies in the span of the coefficient embeddings defining the manifold. We use the form of $\mathbf{w}$ in Eq.~(\ref{spaceproj}) to update the loading vectors in the following step, without explicitly parametrizing the vector $\mathbf{\phi}(\mathbf{c}_{n})$.
\paragraph{\textbf{Trust Region Update for $\{\mathbf{c}_{n}\}$}:}  
The objective function for each patient loading vector $\mathbf{c}_{n}$ decouples as follows when the other variables are fixed:
\begin{multline}
\mathcal{F}(\mathbf{c}_{n}) = \lambda{\vert\vert{\mathbf{y}_{n} - \phi(\mathbf{c}_{n})^{T}\mathbf{w}}\vert\vert}^{2}_{2} + \gamma_{2}{\vert\vert{\mathbf{c}_{n}}\vert\vert}^{2}_{2} + {\Tr\left[{\mathbf{\Lambda}^{T}_{n}}({\mathbf{V}_{n}-\mathbf{X}\mathbf{diag}(\mathbf{c}_{n}))}\right]} \\ +    
{\frac{1}{2}}{\vert\vert{\mathbf{V}_{n}-\mathbf{X}\mathbf{diag}(\mathbf{c}_{n})}\vert\vert}^{2}_{F} \ \ \ \ s.t. \ \ \  \mathbf{c}_{nr} \geq 0
\label{cfcoeff}
\end{multline}
We now substitute this form into Eq.~(\ref{cfcoeff}) and use the kernel trick, to write:
\begin{equation*}
{\vert\vert{\mathbf{y}_{n} - \phi(\mathbf{c}_{n})^{T}\mathbf{w}}\vert\vert}^{2}_{2} = {\vert\vert{\mathbf{y}_{n} - \sum_{j}{\phi(\mathbf{c}_{n})^{T}\phi(\mathbf{\hat{c}_{j}})\boldsymbol{\alpha}_{j}}}\vert\vert_{2}^{2}} = {\vert\vert{\mathbf{y}_{n} - \sum_{j}{\kappa(\mathbf{c}_{n},\mathbf{\hat{c}}_{j})\boldsymbol{\alpha}_{j}}}\vert\vert_{2}^{2}}
\end{equation*}
where $\{\mathbf{{\hat{\mathbf{c}}_{n}}}\}$ denotes the coefficient vector estimates from the previous step to compute $\mathbf{w}$. Notice that the kernel trick buys a second advantage, in that we only need to optimize over the first argument of $\kappa(\cdot,\cdot)$. Since kernel functions typically have a nice analytic form, we can easily compute the gradient ${\nabla\kappa(\mathbf{c}_{n},\mathbf{\hat{c}}_{j})}$ and hessian ${\nabla^{2}\kappa(\mathbf{c}_{n},\mathbf{\hat{c}}_{j})}$ of ${\kappa(\mathbf{c}_{n},\mathbf{\hat{c}}_{j})}$ with respect to $\mathbf{c}_{n}$.
\par Given this, the gradient of $\mathcal{F}(\cdot)$ with respect to $\mathbf{c}_{n}$ takes the following form:
\begin{multline*}
\mathbf{g}_{n} = \frac{\partial{\mathcal{F}}}{\partial{\mathbf{c}_{n}}} = {\mathbf{c}_{n} \circ \left[\left[\mathcal{I}_{R}\circ(\mathbf{X}^{T}\mathbf{X})\right]\mathbf{1}\right]} - \left[\mathcal{I}_{R}\circ(\mathbf{\Lambda}_{n}^{T}\mathbf{X}+\mathbf{V}_{n}^{T}\mathbf{X})\right]\mathbf{1} + 2\gamma_{2}\mathbf{c}_{n} \\ - \lambda{\sum_{i}}{\boldsymbol{\alpha}_{i}}{\left[{2{ \nabla{\kappa(\mathbf{c}_{n},\mathbf{\hat{c}}_{i})}\mathbf{y}_{i} }} - {\sum_{k}}{\boldsymbol{\alpha}_{k}}\left[{\kappa(\mathbf{c}_{n},\mathbf{\hat{c}}_{i})}\nabla{\kappa(\mathbf{c}_{n},\mathbf{\hat{c}}_{k})} + {\kappa(\mathbf{c}_{n},\mathbf{\hat{c}}_{k})}\nabla{\kappa(\mathbf{c}_{n},\mathbf{\hat{c}}_{i})}\right]\right]}
 \label{grad}
\end{multline*}
where $\mathbf{1}$ is the vector of all ones. Notice that the top line of the gradient term is from the matrix decomposition and regularization terms, and the bottom line corresponds to the kernel regression. The Hessian $\mathbf{H}_{n}={\partial^{2}\mathcal{F}}/{\partial{\mathbf{c}_{n}^{2}}}$ can be similarly computed. Due to space limitations, we have omitted its explicit form.
\par Given the low dimensionality of $\mathbf{c}_{n}$, we derive a trust region optimizer for this variable. The trust region algorithm provides guaranteed convergence, like the popular gradient descent method, with the speedup of second-order procedures. The algorithm iteratively updates $\mathbf{c}_{n}$ according to the descent direction $\mathbf{p}_{k}$, i.e. $\mathbf{c}^{(k+1)}_{n} = \mathbf{c}^{(k)}_{n} + \mathbf{p}_{k}$. The vector $\mathbf{p}_{k}$ is computed via the following quadratic objective, which is a second order Taylor expansion of $\mathcal{F}$ around $\mathbf{c}^{k}_{n}$ :
\begin{equation*}
    \mathbf{p} = \argmin_{\mathbf{p}}{\mathcal{F}(\mathbf{c}_{n}^{k})+{\mathbf{g}_{n}^{k}(\mathbf{c}_{n}^{k})^{T}}{\mathbf{p}} + \frac{1}{2}{\mathbf{p}^{T}\mathbf{H}_{n}^{k}(\mathbf{c}_{n}^{k}){\mathbf{p}}}} \ \ s.t. \ {\vert\vert{\mathbf{p}}\vert\vert_{2}\leq \delta_{k} \ , \ \mathbf{c}^{k}_{nr} + \mathbf{p}_{r} \geq 0}
\end{equation*}
where $\mathbf{g}_{n}(\cdot)$ and $\mathbf{H}_{n}(\cdot)$ are the gradient and Hessian referenced above evaluated at the current iterate $\mathbf{c}^{k}_{n}$. We recursively search for a suitable trust region radius $\delta_{k}$ such that we are guaranteed sufficient decrease in the objective at each iteration. This algorithm has a lower bound on the function decrease per update, and with an appropriate choice of the $\mathbf{\delta}_{k}$, converges to a local minimum of $\mathcal{F}$ \cite{wright1999numerical}.

\paragraph{\textbf{Augmented Lagrangian Update for $\mathbf{V}_{n}$ and $\mathbf{\Lambda}_{n}$:}}
Each $\{\mathbf{V}_{n}\}$ has a closed form solution, while the dual variables $\{\mathbf{\Lambda}_{n}\}$ are updated via gradient ascent: 
\begin{eqnarray}
\mathbf{V}_{n} = (\mathbf{diag}(\mathbf{c}_{n})\mathbf{X}^{T}+ 2 \mathbf{\Gamma}_{n}\mathbf{X} - \mathbf{\Lambda}_{n})(\mathcal{I}_{R}+2 \mathbf{X}^{T}\mathbf{X})^{-1} \label{cons}
\\ \mathbf{\Lambda}_{n}^{k+1} = \mathbf{\Lambda}_{n}^{k} + \eta_{k}(\mathbf{V}_{n}-\mathbf{X}\mathbf{diag}(\mathbf{c}_{n})) \ \ \ \ \ \ 
\label{GA}
\end{eqnarray}
We cycle through the updates in Eqs.~(\ref{cons}-\ref{GA}) to ensure that the proximal constraints are satisfied with increasing certainty at each step. We choose the learning rate parameter $\eta_{k}$ for the gradient ascent step of the Augmented Lagrangian to guarantee sufficient decrease for every iteration of alternating minimization.

\paragraph{\textbf{Prediction on unseen data:}}
We use the estimates $\{\mathbf{X}^{*},\mathbf{w}^{*},\{\mathbf{c}_{n}^{*}\}\}$ obtained from the training data to compute the loading vector $\mathbf{\Bar{c}}$ for an unseen patient. We must remove the data term in Eq.~(\ref{obj2}), as the corresponding value of $\mathbf{\Bar{y}}$ is unknown for the new patient. Hence, the kernel terms in the gradient and hessian disappear. We also assume that the conditions for the proximal operator hold with equality; this eliminates the Augmented Lagrangians in the computation. The objective in $\mathbf{\Bar{c}}$ reduces to the following quadratic form:
\begin{equation}
\frac{1}{2}{\mathbf{\Bar{c}}^{T}\mathbf{\Bar{H}}\mathbf{\Bar{c}}} + \mathbf{\Bar{f}}^{T}\mathbf{\Bar{c}} \ \ s.t. \ \ \mathbf{\Bar{A}}\mathbf{\Bar{c}} \leq \mathbf{\Bar{b}}
\label{quadprog}
\end{equation}
Note that the formulation is similar to the trust region update we used previously. For an unseen patient, the parameters from Eq.~(\ref{quadprog}) are:
\begin{eqnarray*}
 \mathbf{\Bar{H}} = 2(\mathbf{X}^{T}\mathbf{X})\circ(\mathbf{X}^{T}\mathbf{X}) + 2\gamma_{2}\mathcal{I}_{R} \ \ \ \ \ \ \ \   \\  \ \ \ \ \ \  \ \ \mathbf{\Bar{f}} = -2\mathcal{I}_{R}\circ(\mathbf{X}^{T}\mathbf{\Gamma}_{n}\mathbf{X})\mathbf{1}; \ \  \mathbf{\Bar{A}} = -\mathcal{I}_{R} \ \  \mathbf{\Bar{b}} = \mathbf{0} \ \ \ \    
\end{eqnarray*} 
The Hessian $\mathbf{\Bar{H}}$ is positive definite, which leads to an efficient quadratic programming solution to Eq.~(\ref{quadprog}).
The severity score for the test patient is estimated by $\mathbf{\Bar{y}}= \mathbf{\phi}({\mathbf{\Bar{c}}})^{T}\mathbf{w}^{*} = \sum_{j}{\kappa(\mathbf{\Bar{c}},\mathbf{c}^{*}_{j})}\boldsymbol{\alpha}^{*}_{j}$, where $\boldsymbol{\alpha}^{*} = \left[\mathbf{\Phi}(\mathbf{C}^{*})^{T} \mathbf{\Phi}(\mathbf{C}^{*})+\frac{\gamma_{3}}{\lambda}\mathcal{I}_{N}\right]^{-1}\mathbf{y}$.
\subsection{Baseline Comparison Methods}
We compare our algorithm with the standard manifold learning pipeline to predict the target severity score. We consider two classes of representation learning techniques motivated from the machine learning and graph theoretic literature. From here, we construct a non-linear regression model similar to our manifold learning term in Eq.~(\ref{obj}). Our five baseline comparisons are as follows: 
\begin{itemize}
\item[1.]{Principal Component Analysis (PCA) on the stacked $\frac{P\times(P-1)}{2}$ correlation coefficients followed by a kernel ridge regression (kRR) on the projections}
\item[2.]{Kernel Principal Principal Component Analysis (kPCA) on the correlation coefficients followed by a kRR on the embeddings}
\item[3.]{Node Degree computation ($D_{N}$) based on the thresholded correlation matrices followed by a kRR on the $P$ node features}
\item[4.]{Betweenness Centrality ($C_{B}$) on the thresholded correlation matrices followed by a kRR on the $P$ node features}
\item[5.]{Decoupled Matrix Decomposition (Eq.(\ref{obj})) and kRR on the loadings $\{\mathbf{c}_{n}\}$}.
\end{itemize}
Baseline $5$ helps us evaluate and quantify the advantage provided by our joint optimization approach as opposed to a pipelined prediction of clinical severity.
\section{Experimental Results:}
\paragraph{\textbf{rs-fMRI Dataset and Preprocessing.}} We validate our method on a cohort of $58$ children with high-functioning ASD (Age: $10.06 \rpm 1.26$, IQ: $110\rpm 14.03$). rs-fMRI scans were acquired on a Phillips $3$T Achieva scanner using a single-shot, partially parallel gradient-recalled EPI sequence with TR/TE $=2500/30$ms, flip angle $=70^{\circ}$, res $=3.05\times3.15\times3$mm, having $128$ or $156$ time samples.
We use a standard pre-processing pipeline, consisting of slice time correction, rigid body realignment, normalization to the EPI version of the MNI template, Comp Corr \cite{behzadi2007component}, nuisance regression, spatial smoothing by a $6$mm FWHM Gaussian kernel, and bandpass filtering between $0.01-0.1$Hz. We use the Automatic Anatomical Labeling (AAL) atlas to define $116$ cortical, subcortical and cerebellar regions. We subtract the contribution of the first eigenvector from the regionwise correlation matrices because it is roughly constant and biases the predictions. The residual correlation matrices, $\{\mathbf{\Gamma}_{n}\}$, are used as inputs for all the methods.
\par
\begin{wrapfigure}[21]{R}{0.30\textwidth}
\small
\begin{center}
 \centerline{\includegraphics[scale =0.42]{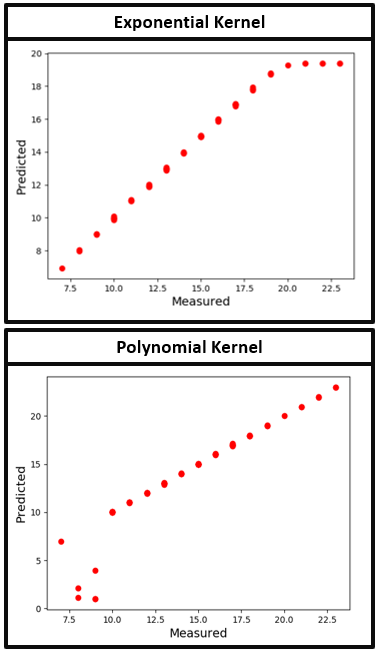}}
 \medskip 
{\caption{\footnotesize{Recovery \textbf{Top:} Exponential \textbf{Bottom:} ~Polynomial Kernel}}\label{Exp_Poly}}
\end{center}
\end{wrapfigure}

 We consider three separate measures of clinical severity quantifying different impairments associated with ASD. The Autism Diagnostic Observation Schedule (ADOS) \cite{payakachat2012autism} captures social and communicative deficits of the patient along with repetitive behaviors (dynamic range: $0-30$).  The Social Responsiveness Scale (SRS) \cite{payakachat2012autism} characterizes impaired social functioning (dynamic range: $70-200$). Finally, the Praxis score \cite{dowell2009associations} quantifies motor control, tool usage and gesture imitation skills in ASD patients (dynamic range: $0-100$).
\paragraph{\textbf{Characterizing the Non-Linear Patient Manifold:}}
Based on simulated data, we observed that the standard exponential kernel provides a good recovery performance in the lower part of the dynamic range, while polynomial kernels are more suited for modeling the larger behavioral scores, as shown in Fig~\ref{Exp_Poly}. Thus, we use a mixture of both kernels to capture the complete behavioral characteristics:
\begin{equation*}
\kappa(\mathbf{c}_{i},\mathbf{c}_{j}) = \mathbf{\exp}\left[-\frac{{\vert\vert{\mathbf{c}_{i}-\mathbf{c}_{j}}\vert\vert}^{2}_{2}}{\sigma^{2}}\right] + \frac{\rho}{l}{\left(\mathbf{c}_{j}^{T}\mathbf{c}_{i} + 1\right)^{l}} 
\end{equation*}
We vary the kernel parameters across $2$ orders of magnitude and select the settings: ADOS~$\{{\sigma}^2=1, \rho = 0.8, l = 2.5\}$, SRS~$\{{\sigma}^2 =1, \rho = 2, l = 1.5\}$ and Praxis~ $\{{\sigma}^2 =1, \rho = 0.5, l = 1.5\}$. The varying polynomial orders reflect the differences in the dynamic ranges of the scores.
\subsubsection{{Predicting ASD Clinical Severity.}}
We evaluate every algorithm in a ten fold cross validation setting, i.e. we train the model on a $90$ percent split of our data, and report the performance on the unseen $10$ percent. The number of components was fixed at $15$ for PCA and at $10$ for k-PCA. For k-PCA, we use an RBF kernel with the coefficient parameter $0.1$. There are two free parameters for the kRR, namely, the kernel parameter $C$ and $\ell_{2}$ parameter $\beta$. We obtain the best performance for the following settings: ADOS~$\{C=0.1,\beta=0.2\}$, SRS~$\{C=0.1,\beta=0.8\}$, and Praxis~$\{C=0.01,\beta=0.2\}$. For the graph theoretic baselines, we obtained the best performance by thresholding the entries of $\{\mathbf{\Gamma}_{n}\}$ at $0.2$. We fixed the parameters in our CMO framework using a grid search for $\{\lambda,\gamma_{1},\gamma_{2},\gamma_{3}\}$. The values were varied between $(10^{-3}- 10)$. The performance is insensitive to $\lambda$ and $\gamma_{3}$, which are fixed at $1$. The remaining parameters were set at $\{\gamma_1=10,\gamma_2=0.7,\gamma_3=1\}$ for all the scores. We fix the number of networks, $R$, at the knee point of the eigenspectrum of $\{\mathbf{\Gamma}_{n}\}$, i.e. $(R=8)$.
\begin{table}[b!]
\footnotesize
\centering
{\caption{\footnotesize{Performance evaluation using \textbf{Median Absolute Error (MAE)} \& \textbf{Mutual Information (MI)}. Lower MAE \& higher MI indicate better performance.}}\label{table:1}}
\begin{tabular}{|c |c | c| c| c| c|} 
\hline 
  \textbf{Score} &\textbf{Method} &\textbf{MAE Train} & \textbf{MAE Test} & \textbf{MI Train} & \textbf{MI Test} \\  
\hline 
\hline
  \multirow{6}{4em}{ADOS} & PCA \& kRR & 1.29 & 3.05 & 1.46 & 0.87\\
 & k-PCA \& kRR & 1.00 & 2.94 & 1.48 & 0.38 \\
 & $C_{B}$ \& kRR & 2.10 & 2.93 & 1.03 & 0.95 \\
 & $D_{N}$ \& kRR & 2.09 & 3.03 & 0.97 & 0.96 \\
 & Decoupled & 2.11 & 3.11 & 0.82 & 1.24\\
 & \textbf{CMO Framework} & \textbf{0.035}& \textbf{2.73}& \textbf{3.79}& \textbf{2.10}\\
[0.2ex]  
\hline
 \multirow{6}{4em}{SRS} & PCA \& kRR & 7.39 & 19.70 & 2.78 & 3.30 \\
 & k-PCA \& kRR & 5.68 & 18.92 & 2.85 & 1.74\\
 & $C_{B}$ \& kRR & 11.00 & 17.72 & 2.32 & 3.66\\
 & $D_{N}$ \& kRR & 11.46 & 17.79 & 2.24 & 3.60\\
 & Decoupled & 15.9 & 18.61 & 2.04 & 3.71\\
 & \textbf{CMO Framework} & \textbf{0.09}& \textbf{13.28}& \textbf{5.28}& \textbf{4.36}
 \\ [0.2ex]
 \hline
 \multirow{6}{4em}{Praxis} & PCA \& kRR & 5.33 & 12.5 & 2.50& 2.68 \\
 & k-PCA \& kRR & 4.56 & 11.15 & 2.56 & 1.51\\
 & $C_{B}$ \& kRR & 8.17 & 12.61 & 1.99 & 3.05\\
 & $D_{N}$ \& kRR & 8.18 & 13.14 & 2.00 & 3.20\\
 & Decoupled & 10.11 & 13.33 & 3.28& 1.53\\
 & \textbf{CMO Framework} & \textbf{0.13}& \textbf{9.07}& \textbf{4.67}& \textbf{3.87}\\
 [0.4ex]
 \hline
\end{tabular}
\end{table}
\begin{figure}[b!]     
    \centering
      \includegraphics[width=\dimexpr \textwidth-2\fboxsep-2\fboxrule\relax, scale=0.65]{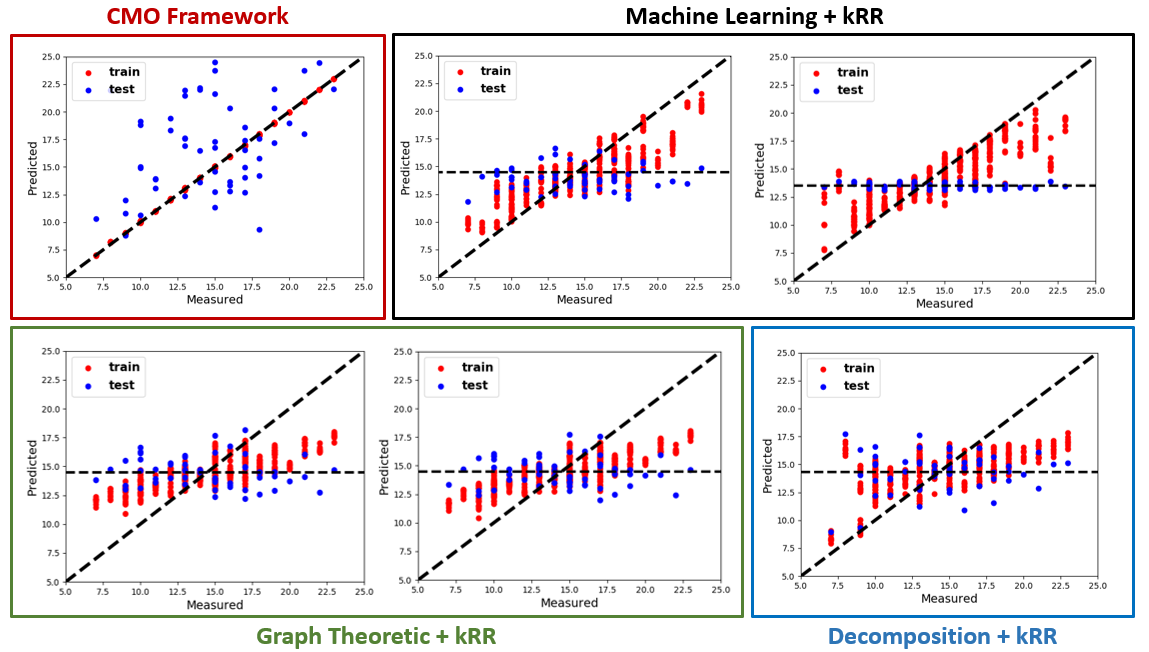}
   {\caption{\footnotesize{Prediction performance for the ADOS score for \textbf{Red Box:} CMO Framework. \textbf{Black Box:} \textbf{(L)} PCA and kRR \textbf{(R)} k-PCA and kRR, \textbf{Green Box:} \textbf{(L)} Node Degree Centrality and kRR \textbf{(R)} Betweenness Centrality and kRR \textbf{Blue Box:} Matrix Decomposition from Eq. (\ref{obj}) followed by kRR}}\label{ADOS}} 
\end{figure}%
\subsubsection{Performance Comparison.}
\begin{figure}[t!]    
    \centering
      \includegraphics[width=\dimexpr \textwidth-2\fboxsep-2\fboxrule\relax, scale=0.65]{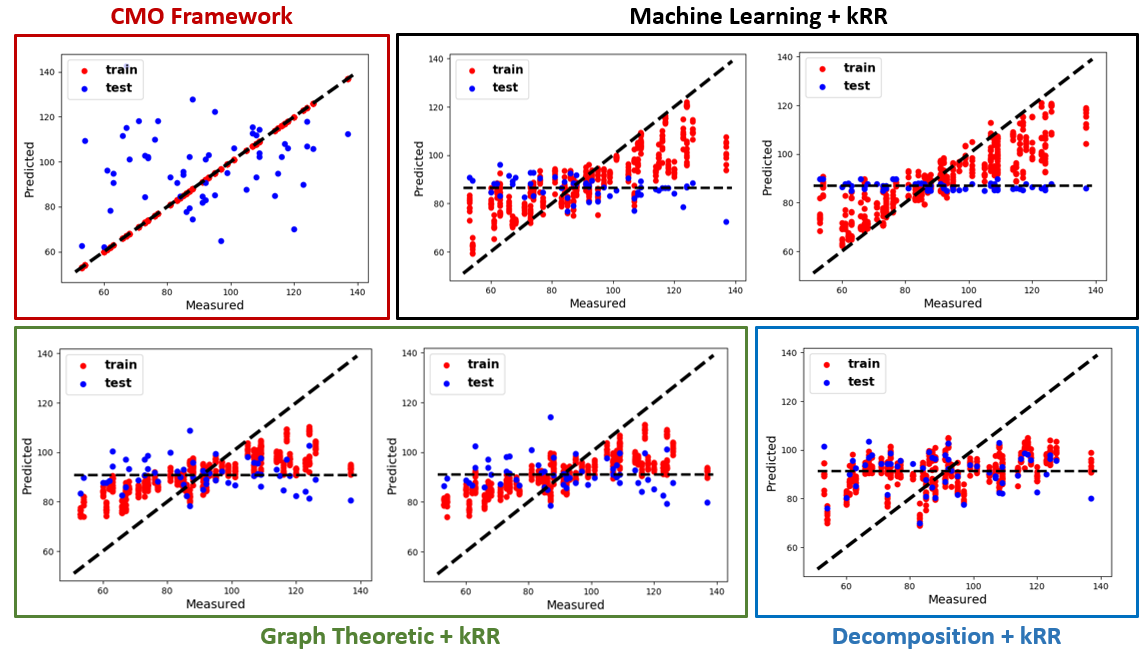}
   \small{\caption{\footnotesize{Prediction performance for the SRS score for \textbf{Red Box:} CMO Framework. \textbf{Black Box:} \textbf{(L)} PCA and kRR \textbf{(R)} k-PCA and kRR, \textbf{Green Box:} \textbf{(L)} Node Degree Centrality and kRR \textbf{(R)} Betweenness Centrality and kRR \textbf{Blue Box:} Matrix Decomposition from Eq. (\ref{obj}) followed by kRR}}\label{SRS}} 
\end{figure}

\begin{figure}[h!]    
    \centering
      \includegraphics[width=\dimexpr \textwidth-2\fboxsep-2\fboxrule\relax, scale=0.65]{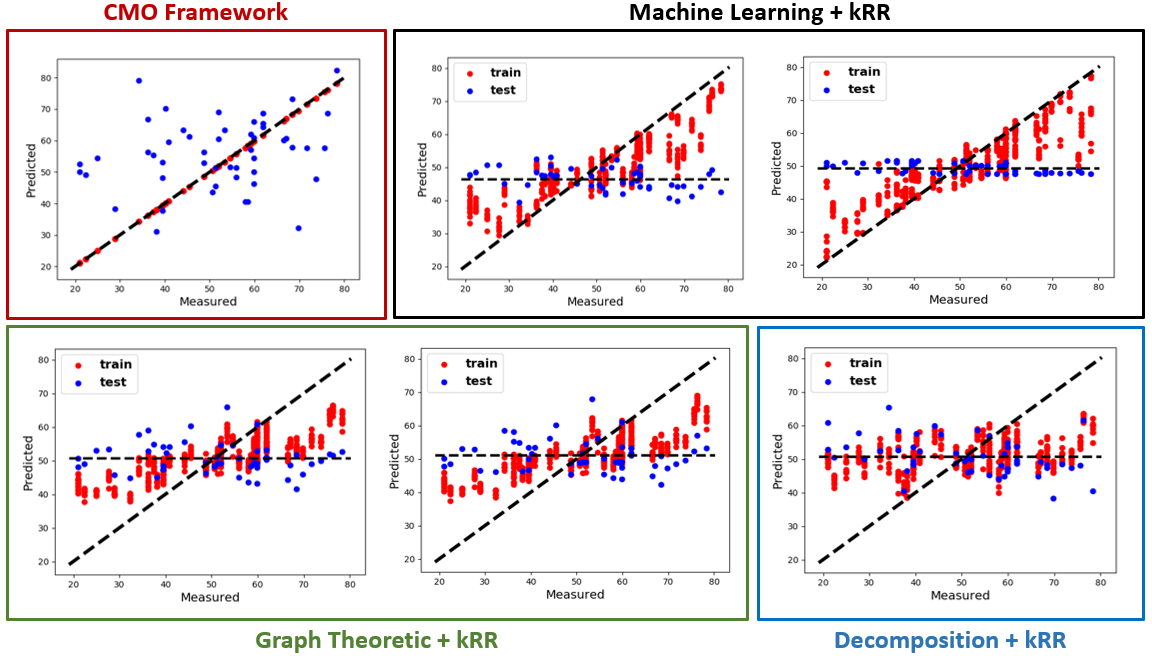}
   {\caption{\footnotesize{Prediction performance for the Praxis score for \textbf{Red Box:} CMO Framework. \textbf{Black Box:} \textbf{(L)} PCA and kRR \textbf{(R)} k-PCA and kRR, \textbf{Green Box:} \textbf{(L)} Node Degree Centrality and kRR \textbf{(R)} Betweenness Centrality and kRR \textbf{Blue Box:} Matrix Decomposition from Eq. (\ref{obj}) followed by kRR}}\label{Praxis}} 
\end{figure}

\begin{figure}[t!]
 \centerline{\includegraphics[scale=0.365]{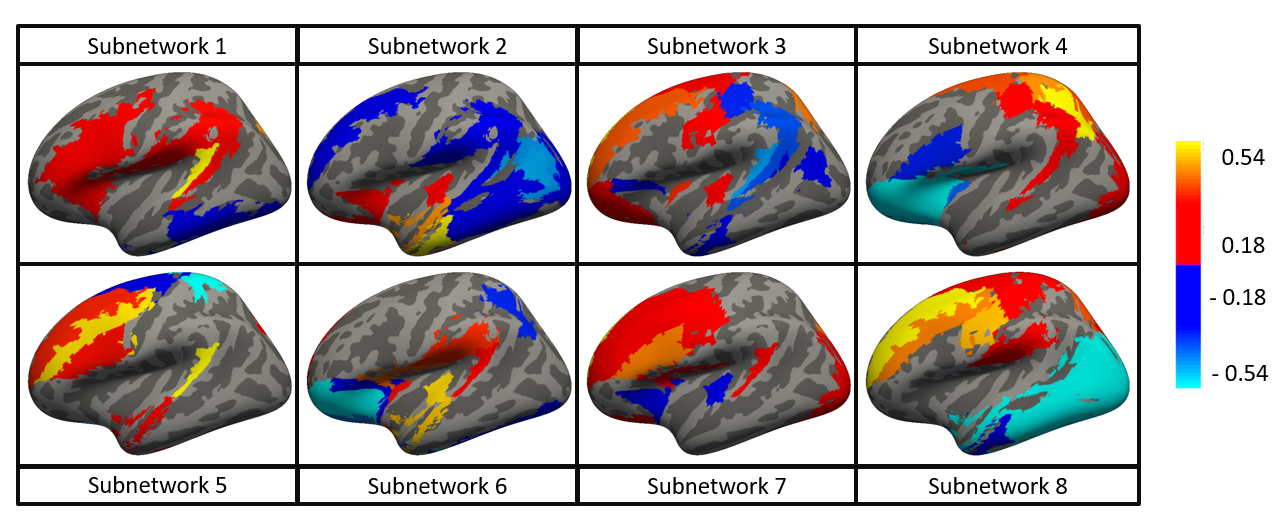}}
 \medskip 
{\caption{\footnotesize{Eight subnetworks identified by our model from the prediction of  ADOS. The blue \& green regions are anticorrelated with the red \& orange regions.}} \label{ADOS_NL}}
\end{figure}

Fig.~\ref{ADOS}, Fig.~\ref{SRS}, and Fig.~\ref{Praxis} illustrate the regression performance for ADOS, SRS, and Praxis respectively. The bold $\mathbf{x}=\mathbf{y}$ line indicates ideal performance. The red points denote the training fit, while the blue points indicate testing performance. Note that baseline testing performance tracks the mean value of the data (indicated by the horizontal black line). In comparison, our method not only consistently fits the training set more faithfully, but also generalizes much better to unseen data. We emphasize that even the pipelined treatment using the matrix decomposition in Eq.~(\ref{obj}), followed by a kernel ridge regression on the learnt projections fails to generalize. This finding makes a strong case for coupling the two representation terms in our CMO strategy. We conjecture that the baselines fail to capture representative connectivity patterns that explain both the functional neuroimaging data space and the patient behavioral heterogeneity. On the other hand, our CMO framework leverages the underlying structure of the correlation matrices through the basis manifold representation. At the same time, it seeks those embedding directions that are predictive of behavior. As reported in Table \ref{table:1}, our method quantitatively outperforms the baselines approaches, in terms of both the Median Absolute Error (MAE) and the Mutual Information (MI) metrics.         
\paragraph{\textbf{Clinical Interpretation.}}
Fig.~\ref{ADOS_NL} illustrates the subnetworks $\{\mathbf{X}_{r}\}$ trained on ADOS. The colorbar indicates subnetwork contributions to the AAL regions. Regions storing negative values are anticorrelated with positive regions. From a clinical standpoint, Subnetwork~$4$ includes the somatomotor network (SMN) and competing i.e. anticorrelated contributions from the default mode network (DMN), previously reported in ASD \cite{nebel2016intrinsic}. Subnetwork~$8$ comprises of the SMN and competing contributions from the higher order visual processing areas in the occipital and temporal lobes. These findings are in line with behavioral reports of reduced visual-motor integration in ASD \cite{nebel2016intrinsic}. Though not evident from the surface plots, Subnetwork~$5$ includes anticorrelated contributions from subcortical regions, mainly, the amygdala and hippocampus, believed to be important for socio-emotional regulation in ASD. Finally, Subnetwork~$6$ has competing contributions from the central executive control network and insula, which are critical for switching between self-referential and goal-directed behavior \cite{sridharan2008critical}. \par
\begin{figure}[b!]
 \centerline{\includegraphics[scale=0.35]{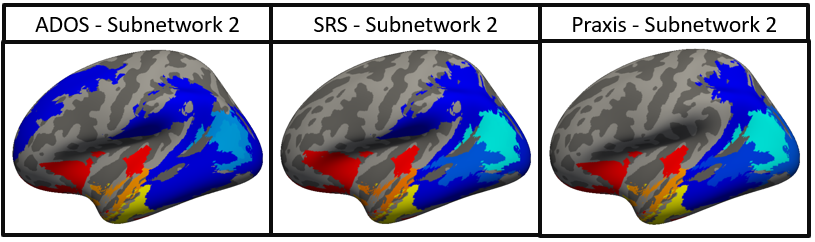}}
 \medskip 
\small{\caption{\footnotesize{Subnetwork $2$ obtained from \textbf{L:} ADOS \textbf{M:} SRS and \textbf{R:} Praxis prediction}} \label{Net_Comp}}
\end{figure}

 Fig.~\ref{Net_Comp} compares Subnetwork~$2$ obtained from ADOS, SRS and Praxis prediction. There is a significant overlap in the bases subnetworks obtained by training across the different scores. This strengthens the hypothesis that our method is able to identify representative, as well as predictive connectivity patterns.
\section{Conclusion}
We have introduced a  Coupled Manifold Optimization strategy that jointly analyzes data from two distinct, but related, domains through its shared projection. In contrast to conventional manifold learning, we optimize for the relevant embedding directions that are predictive of clinical severity. Consequently, our method captures representative connectivity patterns that are important for quantifying and understanding the spectrum of clinical severity among  ASD patients. We would like to point out that our framework makes very few assumptions about the data and can be adapted to work with different similarity matrices and clinical scores. We believe that our method could potentially be an important diagnostic tool for the cognitive assessment of various neuropsychiatric disorders. We are working on a multi-score extension which jointly analyses different behavioral domains. We will explore extensions of our representation that simultaneously integrate functional, structural and behavioral information.
\subsubsection{Acknowledgements.}
This work was supported by the National Science Foundation CRCNS award 1822575, National Science Foundation CAREER award 1845430, the National Institute  of Mental Health (R01 MH085328-09, ~R01 MH078160-07, K01 MH109766 and R01 MH106564), the National Institute of Neurological Disorders and Stroke (R01NS048527-08), and the Autism Speaks foundation.
\bibliographystyle{splncs04}
{\bibliography{MyRefs.bib}
\end{document}